\pdfoutput=1

\documentclass[11pt]{article}
\usepackage[]{ACL2023}
\usepackage{times}
\usepackage{latexsym}
\usepackage[T1]{fontenc}
\usepackage[utf8]{inputenc}
\usepackage{microtype}
\usepackage{inconsolata}
\usepackage{amssymb}
\usepackage{multirow}
\usepackage{comment}
\usepackage{amsmath}
\usepackage{booktabs}
\usepackage{colortbl}
\usepackage{xcolor}
\usepackage{graphicx}

\title{PerPLM: Personalized Fine-tuning of Pretrained Language Models \\ via Writer-specific Intermediate Learning and Prompts}

\author{
{\bf Daisuke Oba} \quad
{\bf Naoki Yoshinaga} \quad
{\bf Masashi Toyoda} \quad\\
Institute of Industrial Science, The University of Tokyo\\
{\tt {oba}@tkl.iis.u-tokyo.ac.jp}\quad
{\tt \{ynaga, toyoda\}@iis.u-tokyo.ac.jp} 
}

\begin{document}
\maketitle

\begin{abstract}
The meanings of words and phrases depend not only on where they are used (contexts) but also on who use them (writers). Pretrained language models (PLMs) are powerful tools for capturing context, but they are typically pretrained and fine-tuned for universal use across different writers. This study aims to improve the accuracy of text understanding tasks by personalizing the fine-tuning of PLMs for specific writers. We focus on a general setting where only the plain text from target writers are available for personalization. To avoid the cost of fine-tuning and storing multiple copies of PLMs for different users, we exhaustively explore using writer-specific prompts to personalize a unified PLM\@. Since the design and evaluation of these prompts is an underdeveloped area, we introduce and compare different types of prompts that are possible in our setting. To maximize the potential of prompt-based personalized fine-tuning, we propose a personalized intermediate learning based on masked language modeling to extract task-independent traits of writers' text. Our experiments, using multiple tasks, datasets, and PLMs, reveal the nature of different prompts and the effectiveness of our intermediate learning approach.
\end{abstract}

\section{Introduction}
{Most of the} natural language processing models assume that our natural language is universal across writers, e.g., word2vec~\cite{mikolov-etal-2013-linguistic} and BERT~\cite{devlin-etal-2019-bert}. However, it has been reported that people handle language in their own way~\cite{lynn-etal-2017-human,oba2019modeling,welch2020exploring}. 
It indicates, for better understanding of text, we need to not only capture linguistic context explicitly obtained from the textual surface, but also consider the {writer} traits of language.

Pretrained language models (PLMs), such as BERT~\cite{devlin-etal-2019-bert} and GPT-3~\cite{brown2020language}, excel {at understanding} the linguistic context by computing contextualized representations of text, and have been a dominant choice of architecture in various text understanding tasks~\cite{li-etal-2019-exploiting,du-etal-2020-adversarial,abuzayed2021sarcasm}.
Recently, researchers have begun trying to equip PLMs with the latter ability to consider the personalization of writers.
\citet{hofmann2021dynamic} constructed and encoded a user graph based on interpersonal features (\textit{e.g.}, follower or not) to personalize BERT\@.
However, not only the rules for graph construction differ across datasets, 
but writer-relationships are not observable in most datasets.
As a data- and task-agnostic approach, \citet{zhong-etal-2021-useradapter} optimized a single length of writer-specific vector which is inserted to the input layer. 
It is similar to, but a simplified version of the Prompt Tuning ~\cite{liu2021p,lester-etal-2021-power}, in which multiple length of task-specific vectors (soft prompts) are inserted in all the layers. Therefore, they may not maximize the full potential of Prompt Tuning techniques.
\citet{mireshghallah-etal-2022-useridentifier} augmented the original input with writer-specific random token sequence (\textit{Hard-Prompts}).
However, random tokens do not have any writer-specific information other than that for distinguishing writers.

This study proposes personalized intermediate learning and fine-tuning of PLMs for more accurate text understanding, in a task- and data-agnostic setting, where no additional resources other than plain text written by target writers can be used for personalization. Both of learning and tuning is based on prompts to exploit a single PLM across writers for the memory efficiency. We first compare different types of personalized soft and hard prompts, including the existing and novel ones, which are used for both fine-tuning and intermediate learning. We propose methods for inserting multiple length of writer-specific soft prompts in all layers to take full advantage of Prompt Tuning techniques~(\S~\ref{sec:method:soft}) and for augmenting inputs using the writers' plain text that have explicit information about ``\textit{what kind of text each writer writes}''~(\S~\ref{sec:method:hard}).

For boosting the personalization performance in the prompts-based fine-tuning, inspired by the success of task-specific pretraining or intermediate learning~\cite{yamada-etal-2020-luke,zhang2020pegasus,guu2020realm,liu2021ner,ri-etal-2022-mluke}, we then propose a personalized intermediate learning method, in which we perform masked language modeling on {plain text written by} target user.
It gives the model writer-specific prior knowledge, i.e., ``what kind of text will this writer write?'', that differs from that of the target task.

In experiments, we verify the different nature of various prompts and the effectiveness of our personalized intermediate learning method, by using multiple tasks (i.e., sentiment analysis and hashtag recommendation), PLMs and datasets~(\S~\ref{sec:experiment}). 
Our key contributions are three-folds; we
\begin{itemize}
    \item propose several methods to achieve personalized text understanding with PLMs based on personalized prompts in a task- and data-agnostic setting,
    \item propose personalized intermediate learning, which can capture the writer-specific and task-agnostic traits of user text for PLM personalization, and   
    \item conduct experiments across multiple tasks, datasets, and PLMs, and provide empirical analysis on the different natures of prompts.
\end{itemize}

\section{Related Work}\label{sec:related_work}
Several existing studies have tried to capture potential personal biases in natural language.
Some of them computed static word embeddings for each writer~\cite{amer2016toward,zeng2017socialized,zeng2018biased,oba2019modeling,oba2020personal,welch2020compositional,welch2020exploring}, in which different embedding is assigned to the same word for different users.
Others have tailored RNNs and LSTMs by preparing some writer-specific parameters to handle the entire input~\cite{levit2015personalization,li2016persona,jaech2018personalized,kolchinski2018representing} or by using the models individually fine-tuned or initialized~\cite{yoon2017efficient,shao2020examination,king2020evaluating}. 
Transformer-based PLMs are not only capable of computing context-dependent meanings for each word, \textit{e.g.}, polysemy, but also more efficient for parallel computing than sequential processing networks, \textit{e.g.}, RNNs. We focus on making these PLMs perform personalized text understanding.

\citet{hofmann2021dynamic} have proposed a method to make PLMs' behavior personalized by simultaneously encoding a user graph constructed based on writer's relationships.
It seems to be reasonable since researchers have successfully personalized models other than Transformer with writer-relationship features (\textit{e.g.}, follower-followee)~\cite{mishra2019abusive,del2019you,li2019encoding}. 
However, such relationships are not always available.
We focus on a setting where no additional resources can be used other than the plain text written by writers.

Several studies have personalized PLMs without additional resources such as user graphs.
Inspired by lightweight fine-tuning, Prompt Tuning~\cite{lester-etal-2021-power,liu2021p}, \citet{zhong-etal-2021-useradapter} inserted a single writer-specific vector (i.e., soft prompts) into the input layer, whereas the original Prompt Tuning inserted multiple prompts in {all} layers. 
Although not for text understanding, \citet{li2022personalized} have also inserted prompts only into the first layer to personalize text generation. 
We verify whether their simplified use of prompt tuning is sufficient by inserting multiple writer-specific soft prompts into all layers.

\citet{mireshghallah-etal-2022-useridentifier} have combined a writer-specific random token sequence with the input as Hard Prompts. 
Unlike the conventional hard prompting, 
in which we have to tailor prompts for the purpose (\textit{e.g.}, ``\textit{this is a [MASK] movie}'' for sentiment analysis on movie reviews), random tokens can be implemented independently from data and tasks.
However, random tokens carry no information other than to distinguish writers. 
We focus on different information about ``\textit{what kind of text the writer writes}'' obtained from the writer's plain text, as another candidate of Hard Prompts.

Some have introduced pretraining or intermediate learning that are specific for target downstream tasks
to get maximum value out of Transformer~\cite{yamada-etal-2020-luke,zhang2020pegasus,guu2020realm,liu2021ner,ri-etal-2022-mluke}.
Inspired by them, we propose a task-agnostic personalized intermediate learning method, in which we perform masked language modeling given the writer via personalized prompts. 
It gives PLMs a kind of task-agnostic knowledge for the writer.

\section{Methods}
In this section, we propose personalized intermediate learning and fine-tuning for PLMs toward the (target) text understanding task, both of which are based on prompts.
We focus on the task-independent settings where only plain text of the target writers can be used as additional resource for achieving the personalization.
In what follows, we first introduce various approaches to prompt-based personalization~(\S~\ref{sec:method:soft} and \S~\ref{sec:method:hard}), which can be used for both intermediate learning and fine-tuning.
We then describe our writer-dependent intermediate learning based on the introduced prompts, which is applied before conducting the fine-tuning on the target task~(\S~\ref{sec:method:inter}).

\subsection{Personalized Soft Prompts}\label{sec:method:soft}
Soft prompts-based methods are the personalized versions of Prefix-tuning~\cite{li2021prefix} and Prompt-tuning~\cite{liu2021p,lester-etal-2021-power}, which updates task-specific soft prompts via target task while freezing parameters in the PLM\@.
Specifically, we prepare soft prompts for each writer, and update them while updating or fixing parameters in the PLMs.

Originally, given an input $x=\{x_{0},...,x_{I-1}\}$, each Transformer layer ${l}$ outputs representations 
$\mathbf{e}_{l}=\{\mathbf{e}_{l}(x_{0}),...,\mathbf{e}_{l}(x_{I-1})\}$, which will be input of the next layer ${l+1}$.
In our method, 
we device the fixed number of writer-specific soft prompts on each layer, and exploit them depending on the writer $u$ of input $x$.
Specifically, these prompts are concatenated with the layer output $e_l$, and used as the extended input for the next layer $l+1$:
\begin{align}
  \mathbf{e}_{l}=\{\mathbf{p}_{u}^{l_0},...,\mathbf{p}_{u}^{l_{M-1}},\mathbf{e}_{l}(x_{0}),...,\mathbf{e}_{l}(x_{I-1})\},  
\end{align}
where $\mathbf{p}_{u}^{l_m}\in\mathbb{R}^{H}$ refers to the $m$-th prompts for user $u\in{U}$ in the layer ${l}$.
Each prompt $\mathbf{p}_{u}^{l_m}$ and token representation $\mathbf{e}_l^{x_n}$ have the same dimension $H$\@.
The outputs of the final layer $L$ corresponding to prompts are discarded in the target task.

Following \citet{liu2021p}, we re-parametarize the soft prompts for the improved training speed, robustness, and task performance.
Let $\mathbf{P}\in{\mathbb{R}^{(L\times{M}\times|U|)\times{H}}}$ be the prompt matrix, in which a prompt $p_u^{l_m}$ corresponds to a row.
We define the continuous prompts matrix $\mathbf{P}=$\textsc{MLP}$(\mathbf{P}')$ with a multi-layer perceptron (\textsc{MLP}) and a smaller matrix $\mathbf{P}'\in{\mathbb{R}^{(L\times{M}\times|U|)\times{H'}}}$ where $H'\ll{H}$.
The MLP is shared across writers.
We can save only the transformed $\mathbf{P}$.

We prepare two scenarios for treating parameters in the pretrained model during the optimization;
\textit{fixing} or \textit{updating}.
In writer-agnostic soft prompt tuning~\cite{liu2021p,lester-etal-2021-power}, the pretrained parameters are fixed for computational efficiency.
However, since we prepare prompts for each user, the amount of available training data per prompts becomes relatively sparse.
We evaluate whether freezing PLMs is also applicable in the context of personalization.

\subsection{Personalized Hard Prompts}\label{sec:method:hard}
In this method, we simply augment the input $x$ by using the writer’s plain text as the additional contexts, without making any change to the PLM. Therefor, as for model optimization, all parameters are subject to update as in the normal fine-tuning.
Soft prompts~(\S~\ref{sec:method:soft}) are continuous representations that are induced via the (target) downstream task, whereas this method exploits the set of discrete tokens, hard prompts, with information about what style of language the writer uses.

Here, let $C_u$ be the set of available text for writer $u$. 
In this method, for each input $x$ written by $u$, 
we simply concatenate $x$ and the text in ${C_u}$:
\begin{align}
    x=\{x_{0}, ..., x_{I-1}, {{c}_{u}^{0}}, {{c}_{u}^{1}}, ...\}\\
    {c}_{u}^{j}=\{x^{uj}_0, ...\},
\end{align}
where $c_u^j$ is a text in ${C_u}$.

In practice, all text cannot be incorporated as additional context because of the limited length of the input tokens, $N_{max}$, for each PLM (\textit{e.g.}, $512$ for BERT). 
In this study, we use the fixed number of first $V$ tokens of text as an additional context, and then, we truncate the end of the original input $x$ so that the extended input does not exceed $N_{max}$.

In terms of how to select and feed user text as additional contexts, we have prepared several scenarios: \textit{Hard Prompts}$_{\textsc{static}}$ and \textit{Hard Prompts}$_{\textsc{dynamic}}$.
In \textit{Hard Prompts}$_{\textsc{static}}$, we arrange the text $\in{C_u}$ in a random order, and use the first $V$ tokens as additional context consistently across all the samples of writer $u$. 
By using consistent context for all samples, 
it is expected that the model can easily distinguish between the writers of each text, while referring to the way the writer uses the language.
In \textit{Hard Prompts}$_{\textsc{dynamic}}$, we arrange the text in ${C_u}$ in the order of semantic similarity to the original input $x$, and use the head $V$ token as additional context for $x$.
By dynamically showing the model with the associated context, we expect the limited length of additional context to be of greater significance.

\subsection{Personalized Intermediate Learning}\label{sec:method:inter}
To boost the personalization performance on the subsequent fine-tuning on the target task, we propose a writer-aware, 
task-independent intermediate learning.
Specifically, it conducts masked language modeling task (\textsc{MLM}) with the writer given based on the personalized prompts. It optimizes the model to be able to predict information about ``\textit{what kind of text the writer would write}.'' Meanwhile, personalized fine-tuning~(\S~\ref{sec:method:soft} and \S~\ref{sec:method:hard}) optimizes to predict a task-dependent information: ``\textit{how the writer would perceive (annotate) this text}.''
We expect the intermediate learning can work as prior knowledge for the target task.

We first prepare writer-aware prompts identical to these used for fine-tuning PLM via downstream task.
Then, as in the original MLM, we replace each input token by [MASK] with some probability, and feed their outputs to the classification head to predict the original token. We then calculate the cross-entropy loss to update the parameters.
In \textit{Soft Prompts}~(\S~\ref{sec:method:soft}), we fix the parameters of the PLM\@.
In \textit{Hard Prompts}, the additional context can be masked by the same probability as well, to make use the information in the opposite direction, i.e., ``\textit{who is the likely writer of text} $x$.''

\section{Experiments}\label{sec:experiment}
We evaluate our personalization methods using different tasks and data for text understanding. 
Specifically, we perform sentiment analysis based on Yelp and Amazon review datasets~(\S~\ref{sec:experiment:sentiment}), and hashtag recommendation based on social media posts on Twitter~(\S~\ref{sec:experiment:hashtag}).
From the perspective of computational costs, 
we basically conduct experiments using only BERT-base~\cite{devlin-etal-2019-bert} as the PLM to be personalized. 
Only for sentiment analysis using Yelp dataset, 
we exploit other PLMs, RoBERTa-base~\cite{liu2019roberta} and ELECTRA-base discriminator~\cite{clark2020electra}, to check whether the observed trends from the experiments are architecture-independent.

In the following content, we first describe the baseline model~(\S~\ref{sec:experiment:baseline}), followed by the settings and results for each task~(\S~\ref{sec:experiment:sentiment} and \S~\ref{sec:experiment:hashtag}).

\begin{table}[t]
    \small
    \centering
    \begin{tabular}{lrrr}
    \toprule
    & \textbf{Yelp} & \textbf{Amazon} & \textbf{Twitter}\\
    \midrule
    \textbf{\# users} & 500 & 400 & 1,000 \\
    \textbf{\# samples} & 83,982 & 76,301 & 241,631 \\
    \textbf{\# hashtags} & n/a & n/a & 26,700 \\
    \textbf{avg. \# words} & 168.6 & 304.3 & 10.5 \\
    \bottomrule
    \end{tabular} 
    \caption{Dataset details for sentiment analysis (Yelp and Amazon) and for hashtag recommendation (Twitter).}
    \label{tab:dataset}
\end{table}

\begin{table*}[t]
    \centering
    \small
    \begin{tabular}{lcccccc}
    \toprule

    & {\textbf{Inter}} & \multicolumn{3}{c}{\textbf{Yelp}} & & \textbf{Amazon} \\\cline{3-5}\cline{7-7}
    & & \scriptsize{\textbf{BERT-base}} & \scriptsize{\textbf{RoBERTa-base}} & \scriptsize{\textbf{ELECTRA-base}} & & \scriptsize{\textbf{BERT-base}} \\
    \midrule
    UserIdentifier & &  77.60$\pm$0.23 & 80.08$\pm$0.15 & 79.89$\pm$0.11 & & 75.27$\pm$0.21 \\
    \rowcolor[gray]{0.85} UserIdentifier & $\checkmark$ & 78.82{$\pm$}0.17 & 81.10$\pm$0.22 & 80.78$\pm$0.13 & & 77.28$\pm$0.07 \\
    UserAdapter & &  76.09$\pm$0.21 & 79.02$\pm$0.19& 78.52$\pm$0.10 & & 74.06$\pm$0.30 \\
    \rowcolor[gray]{0.85} UserAdapter & $\checkmark$ & 77.88$\pm$0.13 & 79.29$\pm$0.10 & 79.20$\pm$0.20& & 75.28$\pm$0.14 \\
    Soft-Prompts$_{\textsc{fix}}$ & & 75.33$\pm$0.15 & 79.14$\pm$0.14 & 79.04{$\pm$}0.39 & &  72.78$\pm$0.27 \\
    \rowcolor[gray]{0.85} Soft-Prompts$_{\textsc{fix}}$ & $\checkmark$ & 76.28$\pm$0.09 & 79.75$\pm$0.14 & 80.14$\pm$0.11 & & 72.67$\pm$0.21 \\
    Soft-Prompts & & {77.34}$\pm$0.15 & 78.82$\pm$0.12 & 79.59$\pm$0.17 & & 74.81$\pm$0.23 \\
    \rowcolor[gray]{0.85} Soft-Prompts & $\checkmark$ & 78.76{$\pm$}0.23 & 79.67$\pm$0.34 & 80.40$\pm$0.22 & & 76.09{$\pm$}0.24 \\
    Hard-Prompts$_{\textsc{static}}$ & & 77.18$\pm$0.25 & 79.39$\pm$0.10 & 79.49$\pm$0.23 & & 74.49$\pm$0.55 \\
    \rowcolor[gray]{0.85} Hard-Prompts$_{\textsc{static}}$ & $\checkmark$ &  78.60{$\pm$}0.13 & 81.01$\pm$0.19 & 80.74$\pm$0.07 & & 77.05$\pm$0.09 \\
    Hard-Prompts$_{\textsc{dynamic}}$ & & 75.99$\pm$0.21 & 78.88$\pm$0.20 & 78.36$\pm$0.24 & & 73.67$\pm$0.29 \\
    \rowcolor[gray]{0.85} Hard-Prompts$_{\textsc{dynamic}}$ & $\checkmark$ & 77.06$\pm$0.21 & 79.22$\pm$0.21 & 79.10$\pm$0.10 & & 75.67$\pm$0.26 \\
    \hline
    Fine Tuning \scriptsize & & 76.31$\pm$0.14 & 78.92$\pm$0.31 & 78.22{$\pm$}0.19& & 74.01$\pm$0.36 \\
    \hline
    \end{tabular}
    \caption{Results for sentiment analysis. \textbf{Inter} refers to the personalized intermediate learning. We run experiments five times with different seeds, and show the average performance and standard deviation.}
    \label{tab:results-sentiment}
\end{table*}

\subsection{Baseline Methods}\label{sec:experiment:baseline}
\paragraph{Fine Tuning:}
This method, the normal fine-tuning, does not take the writer of the text into account. 
It does not extend the input, nor does it equip the model with additional parameters. 
All parameters are subject to update.
By comparing to this writer-agnostic baseline, we can observe the degree to which each method succeeds in personalizing the PLM for the target task.

\paragraph{UserAdapter~\cite{zhong-etal-2021-useradapter}:}
It is a simplified version of our Soft-Prompts method~(\S~\ref{sec:method:soft}). 
Specifically, for input $x$ written by user $u$, 
they prepare only a single soft prompt for each user $u$, $p_u$, and insert it only in the first Transformer layer. 
Meanwhile, our Soft-Prompts method inserts $M (\geq 1)$ soft-prompts into every layer for each user, following existing studies using Prompt Tuning for non-personalization purpose~\cite{liu2021p,li2021prefix}.
Our insertion of writer-specific soft prompts besides the input layer is also motivated by the report by \citet{liu2021p} that soft prompts in the layers close to the final layer have larger influence on task performance.
With this baseline, we verify whether it is sufficient to have single-length prompts for each user and to insert it into the input layer.

\paragraph{UserIdentifier~\cite{mireshghallah-etal-2022-useridentifier}:}
This method provides a random token sequence for each user $u$ and combines them with each input $x$. 
It can be regarded as a different version of our \textit{Hard-Prompts}~(\S~\ref{sec:method:hard}).
Although both methods may distinguish each writer, random tokens have no information about what kind of text the writer writes. With this baseline, we verify whether the writer's historical text has more information than that only distinguishing writers.

\subsection{Sentiment Analysis}\label{sec:experiment:sentiment}
\subsubsection{Settings}
\paragraph{Task definition}
We conduct sentiment analysis task as a multiclass classification problem, in which we classify each text into three classes: \texttt{POSITIVE}, \texttt{NEGATIVE}, and \texttt{NEUTRAL}.
We report macro-F$_1$.

\paragraph{Datasets}
We exploit two kinds of review datasets, Yelp restaurant reviews\footnote{https://www.yelp.com/dataset} and Amazon 5-core review subsets for Movie and TV categories,\footnote{https://jmcauley.ucsd.edu/data/amazon/} which are publicly available for academic use. Each review is annotated with five-star ratings by the writer.
We group reviews with the rating 1-2 into \texttt{NEGATIVE}, reviews with the rating 4-5 into \texttt{POSITIVE}, and the remaining into \texttt{NEUTRAL}.
Unlike hashtag recommendation~(\S~\ref{sec:experiment:hashtag}), which has over ten thousands of classes, the distribution of labels in the sentiment analysis task, which has only three classes, is likely to be skewed. 
Since our focus is not on the label biases, we conduct down-sampling per user, in which we randomly reduced the samples so that the number of samples for each class would be the same.
We then select 500/400 reviewers for Yelp/Amazon reviews, who have at least $25$ examples. 
We then randomly divide reviews into train/dev/test sets per reviewer in the ratio of 8:1:1, and aggregate them into the unified train/dev/test sets.
We call the two datasets \textbf{Yelp} and \textbf{Amazon} respectively~(Table~\ref{tab:dataset}).

\paragraph{Modeling Details}
We feed the output of the PLMs corresponding to the first token of the input $x$ (\textit{e.g.}, \texttt{[CLS]} for BERT) into the linear classifier, and minimize cross-entropy in the training with single Quadro RTX 8000. 
We conduct each experiment five times with different random seeds, and report the average performance and standard deviation for the main results.

\paragraph{Other Hyper-parameters}
We initialize PLMs with that provided by Huggingface (ver. 4.17.0), and the other parameters randomly.
We use batch size 16. 
Soft prompts in each layer and hard prompts, including \textit{UserIdentifier}, has the length of 16 tokens. 
For each method, we search the best learning rate among \{1e-4, 1e-5, 5e-5\}, epochs among \{5, 10, 15\} based on macro-F$_1$ on dev. set.

\subsubsection{Results}
Table~\ref{tab:results-sentiment} shows the results on sentiment analysis, from which we have several observations.

First, most of the methods outperform the non-personalized fine-tuning, while \textit{UserAdapter}, and \textit{Soft-Prompts}$_{\textsc{fix}}$, \textit{Hard-Prompts}$_{\textsc{dynamic}}$ do not confirm the effect of personalization.
It indicates that a single prompt is insufficient.
In addition, task-specific information must be included in the PLM even if multiple prompts are provided at each layer.
Besides, the user's relevant text are not effective for sentiment analysis.

Second, there is no significant difference in performance between UserIdentifier, \textit{Soft-Prompts}, and \textit{Hard-Prompts}$_{\textsc{static}}$.
It suggests the writer's historical text and soft prompts, even if we have multiple ones in all layers, have no more information than that for distinguishing each writer.

Third, our personalized intermediate learning improves the performance for all methods consistently. 
It seems to be effective to obtain information that is different from, but related to the target task, i.e., ``\textit{what kind of text will this writer write}.''

Fourth, in the Yelp dataset, we obtain roughly similar trends over the different PLMs.
It indicates that the personalized methods are independent of the type of PLM to some extent.

\subsection{Hashtag Recommendation}\label{sec:experiment:hashtag}
\subsubsection{Settings}
\paragraph{Task Definition}
We conduct hashtag recommendation as a ranking problem, 
in which we rank hashtags given a tweet.
We adopt 
standard metrics for recommendation, i.e.,
normalized Discounted Cumulative Gain at N (nDCG@N).
In our experiment, we adopt N$=\{5, 10\}$.

\paragraph{Datasets}
For simplicity of evaluation, by using Twitter API, which is publicly available for academic use, we exploit English tweets (excluding mentions, retweets, quotes) with a single hashtag at the end of the tweet, and we hereafter call them ``\textit{tweets}.'' 
We randomly select 1,000 users who have posted at least 100 tweets in 2019 (12 months), and extract their 241,631 tweets and 26,700 unique hashtags.
We remove any URL and emojis from all the tweets. For each user, we select the oldest 80\% tweets as the training set, the most recent 10\% tweets as the test set, and the remaining as the validation set, following the general setup of recommendation task. We then aggregate them into the unified train/dev/test set.

\paragraph{Modeling Details}
We personalize the PLM so that the encoded result of the tweet is close to that of the correct hashtag. 
Specifically, we encode each tweet $x$ and annotated correct hashtag $\hat{x}_{tag}$ by using the same PLM, get the output corresponding to the first token of them ({[CLS]}), and maximize their inner product. 
Assuming that the hashtags' meanings are universal to writers, 
we apply writer-dependent methods (\textit{e.g.}, Soft Prompts~(\S~\ref{sec:method:soft})) only when encoding tweets. 
For stable training, we adopt negative sampling, which is used in ranking-based recommendation studies~\cite{wu2019npa}. 
First, we additionally compute the inner product between $K$-negative hashtags for each tweet $x$ to get the $K+1$ inner product values in total. 
We then apply softmax function to these values to obtain normalized scores for $K+1$ pairs.
In training, we maximize the log-likelihood of the normalized score corresponding to the correct hashtag with single Quadro RTX 8000. 
In evaluation, exact computation of normalized scores between vast volume of tweets and thousands of hashtags is computationally expensive. 
We therefore perform ranking among the correct hashtag and K-negative hashtags, following~\cite{wu2019npa}.
We adopt K =$\{10, 200\}$ for training and evaluation respectively.

\paragraph{Other Hyper-parameters}
We initialize BERT-base with that of Huggingface, 
and other model parameters randomly.
We use batch size of 16 for all datasets. 
Soft prompts in each layer and hard prompts (including UserIdentifier) have the length of 16 tokens. 
For each method, we search the best learning rate among \{1e-4, 1e-5, 5e-5\}, epochs among \{10, 15\} based on nDCG@5 of dev. set. 
\begin{table}[t]
    \centering
    \small 
    \begin{tabular}{l@{\,\,\,}c@{\,\,\,}c@{\,\,\,\,}c}
    \toprule
    {\textbf{BERT-base}} & \textbf{Inter} &  \multicolumn{1}{c}{\textbf{nDCG@5}} & {\textbf{nDCG@10}}  \\
    \midrule
    UserIdentifier  &  & 72.16{$\pm$}0.17 & 73.65{$\pm$}0.17 \\
    \rowcolor[gray]{0.85} UserIdentifier & $\checkmark$  & 72.80{$\pm$}0.14 & 74.33{$\pm$}0.14 \\
    UserAdapter & & 71.11$\pm$0.12 & 72.58$\pm$0.14 \\
    \rowcolor[gray]{0.85} UserAdapter & $\checkmark$ & 69.99$\pm$0.18 & 71.57$\pm$0.11  \\
    Soft-Prompts$_{\textsc{fix}}$ & & 58.03$\pm$0.12 & 59.43$\pm$0.11  \\
    \rowcolor[gray]{0.85} Soft-Prompts$_{\textsc{fix}}$ & $\checkmark$ & 55.88{$\pm$}0.07 & 57.49{$\pm$}0.07\\
    Soft-Prompts & & 72.82$\pm$0.25 & 74.16$\pm$0.23 \\
    \rowcolor[gray]{0.85} Soft-Prompts & $\checkmark$ & 72.33{$\pm$}0.28 & 73.74{$\pm$}0.26  \\ 
    Hard-Prompts$_{\textsc{static}}$ & & 72.59{$\pm$}0.17 & 74.07{$\pm$}0.17 \\
    \rowcolor[gray]{0.85} Hard-Prompts$_{\textsc{static}}$ & $\checkmark$ & 73.11{$\pm$}0.23 & 74.66{$\pm$}0.19 \\
    Hard-Prompts$_{\textsc{dynamic}}$ & & 64.80{$\pm$}0.09 & 66.59{$\pm$}0.11 \\   
    \rowcolor[gray]{0.85} Hard-Prompts$_{\textsc{dynamic}}$ & $\checkmark$ & 65.69{$\pm$}0.11 & 67.47{$\pm$}0.09 \\
    \hline
    Fine Tuning  &  & 60.83{$\pm$0.17}  & 62.78{$\pm$0.16}  \\
    \hline
    \end{tabular}
    \caption{Results for hashtag recommendation. \textbf{Inter} refers to the personalized intermediate learning. We run experiments five times with different seeds, and show the average performance and standard deviation.}
    \label{tab:hashtag-main}
\end{table}

\subsubsection{Results}
Table~\ref{tab:hashtag-main} shows the results for hashtag recommendation, from which we can find several insights.

First, similar to the trend in the sentiment analysis,
almost all models outperform the writer-independent fine-tuning, excluding \textit{Soft-Prompts}$_{\textsc{fix}}$.
In this experiment, since hashtags are encoded in a writer-independent encoding manner using the same PLM~(\S~\ref{sec:experiment}),
\textit{Soft-Prompts}$_{\textsc{fix}}$ cannot learn any task-specific information to encode hashtags, which may explain its lower performance than fine-tuning.

Second, in contrast to the tendency in the sentiment analysis, there is no difference between \textit{Soft-Prompts} and \textit{Hard-Prompts}$_{\textsc{static}}$, but they outperform UserIdentifier.
It indicates that, in hashtag recommendation, the users' historical text and soft prompts induced via target task are more valuable than the user identifiers consisting of random tokens.

Third, although \textit{Hard-Prompts}$_{\textsc{dynamic}}$ is inferior to \textit{Hard-Prompts}$_{\textsc{static}}$ as in the case of sentiment analysis~(\S~\ref{sec:experiment:sentiment}), it outperforms the normal fine-tuning.
Compared to sentiment analysis, where each text is assigned to one of three classes (\texttt{POSITIVE}, \texttt{NEGATIVE}, \texttt{NEUTRAL}), hashtag recommendation has tens of thousands of classes. 
Thus, even if the label class of the retrieved similar text and that of the original input text $x$ are different, there is a room to have more related information between each other's labels in hashtag recommendation, resulting 
 in its success of personalization.

Fourth, as in the sentiment analysis, \textit{UserAdapter} is inferior to \textit{Soft-Prompts} across all metrics.
It indicates the importance of multiple prompts in multiple layers for each writer.

Lastly, our personalized intermediate learning (\textbf{Inter} column) boost the performance of \textit{Hard-Prompts} methods, while it has the opposite effects for the \textit{Soft-Prompts} methods. 
PLMs optimized with \textit{Soft-Prompts} may be over-fitting to intermediate learning task, probably due to both the relatively large volume of learnable parameters ranging all layers and the short length of text in Twitter dataset~(Table~\ref{tab:dataset}).

\begin{table}[t]
    \small
    \centering
    \begin{tabular}{l@{\,\,\,\,}c@{\,\,\,}c@{\,\,\,}c@{\,\,\,}c}
    \toprule
    \multirow{2}{*}{\textbf{BERT-base}} & \multirow{2}{*}{\textbf{{Inter}}} & \textbf{{Yelp}} & \textbf{{Amazon}} & \textbf{{Twitter}}\\
     & & \scriptsize{\textbf{{macro-F1}}} & \scriptsize{\textbf{{macro-F1}}} & \scriptsize{\textbf{{nDCG@5}}}\\
    \midrule
    UserIdentifier   & & 76.94 & 76.23 & 71.27 \\
    \rowcolor[gray]{0.85}UserIdentifier   & $\checkmark$ & 77.68 & 78.03 & 72.73 \\
    UserAdapter      & &74.85 & 73.79 & 70.34 \\
    \rowcolor[gray]{0.85}UserAdapter      & $\checkmark$ & 76.40 & 74.34 & 70.32 \\
    Soft-Prompts$_{\textsc{fix}}$ & &73.59 & 72.47 & 52.41 \\ 
    \rowcolor[gray]{0.85}Soft-Prompts$_{\textsc{fix}}$ & $\checkmark$ & 73.17 & 71.57 & 50.90 \\ 
    Soft-Prompts & &75.75 & 75.33 & 71.82 \\
    \rowcolor[gray]{0.85}Soft-Prompts & $\checkmark$ & 77.30 & 77.46 & 72.72 \\
    Hard-Prompts$_{\textsc{static}}$ & &76.20 & 75.33 & 71.24 \\
    \rowcolor[gray]{0.85}Hard-Prompts$_{\textsc{static}}$ & $\checkmark$ & 77.19 & 77.90 & 72.48\\
    Hard-Prompts$_{\textsc{dynamic}}$ & & 74.86 & 72.86 & 66.31 \\
    \rowcolor[gray]{0.85}Hard-Prompts$_{\textsc{dynamic}}$ & $\checkmark$ & 75.85& 76.81 & 67.18 \\
    \hline
    %Fine-tuning (FT) & & 74.40 & 74.22 & 62.54 \\
    Writer-wise FT    & & 74.58 & 73.84 & 69.07 \\
    \hline
    \end{tabular}
    \caption{The performace for the randomly selected 50 writers. \textbf{Writer-wise FT} refers to the writer-wise fine-tuning approach, in which we fine-tune the PLMs for each writer. We run experiments five times with different seeds, and show the average performance.}
    \label{tab:one-by-one}
\end{table}

\section{Analysis}
In this section, we analyze our personalization methods through additional experiments.

\subsection{Comparison to Writer-wise Fine-tuning}
We verify the usefulness of personalized fine-tuning of PLM for each person.
Although the most strait-forward approach is writer-dependent fine-tuning, 
it has not been explored in the existing studies~\cite{zhong-etal-2021-useradapter,mireshghallah-etal-2022-useridentifier}. 
This is probably due to the cost of optimizing and saving different copies of the big PLMs for different users.
To perform evaluation under the realistic computational costs, we randomly select 50 writers for each dataset and perform optimization for each writer independently.
The search space of hyper-parameters is the same with our main experiments~(\S~\ref{sec:experiment}). 
Since data sparseness cannot be ignored if only one writer's data is available, we exploit the parameters of the \textit{Fine-tuning} (\S~\ref{sec:experiment:baseline}), which is optimized with all the writers' data, as the initialization. 
We aggregate the inference results for each writer, and report average macro-$_1$ for sentiment analysis and nDCG@5 for hashtag recommendation. 
We conduct experiments five times independently and report the average performance.

Table~\ref{tab:one-by-one} summarizes the results.
The writer-wise Fine-tuning approach performs worse than most of other scalable approaches across tasks and datasets, even though prompt-based methods are approximately $|U|$ times more memory-efficient.
The result suggests the usefulness of not only treating individual data differently, but also providing the model to them in an integrated manner.
The lack of performance improvement from Fine-tuning on the Yelp and Amazon also suggests the difficulty to make use of personal data independently.

Nevertheless, writer-wise fine-tuning consistently outperforms \textit{Soft-Prompts}$_{\textsc{fix}}$. 
It demonstrates the needs for updating the PLMs themselves for the target task (\S~\ref{sec:experiment}).

\begin{table}[t]
    \small
    \centering
    \begin{tabular}{l@{\,\,\,}c@{\,}r@{\,\,\,\,}r@{\,\,\,\,}r@{\,\,\,\,}r@{\,\,\,\,}r@{\,\,\,\,}r}
    \toprule
    \multirow{2}{*}{\textbf{BERT-base}} & \multirow{2}{*}{{\textbf{{Inter}}}} & \multicolumn{2}{c}{\textbf{{Yelp}}} & \multicolumn{2}{c}{\textbf{{Amazon}}} & \multicolumn{2}{c}{\textbf{{Twitter}}}\\
    & & \scriptsize{\textbf{{\textit{Ga}$\uparrow$}}} &  \scriptsize{\textbf{{\textit{Gb}$\downarrow$}}} &  \scriptsize{\textbf{{\textit{Ga}$\uparrow$}}} & \scriptsize{\textbf{{\textit{Gb}$\downarrow$}}} & \scriptsize{\textbf{{\textit{Ga}$\uparrow$}}} & \scriptsize{\textbf{{\textit{Gb}$\downarrow$}}} \\
    \midrule
    UserIdentifier   & & 18.4 & 9.2 & 5.0 & 2.8 & 64.5 & 2.5 \\
    \rowcolor[gray]{0.85}UserIdentifier   & $\checkmark$ & 22.4 & 6.0 & 22.3 & 6.3 & 66.8 & 2.4 \\
    UserAdapter      & & 2.6 & 4.2 & 9.5 & 9.5 & 57.4 & 2.0 \\
    \rowcolor[gray]{0.85}UserAdapter      & $\checkmark$ & 18.2 & 9.4 & 18.3 & 10.3 & 55.0 & 3.1 \\
    Soft-Prompts$_{\textsc{fix}}$ & & 23.6 & 26.6 & 15.0 & 23.3 & 33.3 & 33.9 \\ 
    \rowcolor[gray]{0.85}Soft-Prompts$_{\textsc{fix}}$ & $\checkmark$ & 23.8 & 22.8 & 15.3 & 18.3 & 29.5 & 39.1 \\ 
    Soft-Prompts & & 15.8 & 9.4 & 4.5 & 2.8 & 64.0 & 1.5 \\
    \rowcolor[gray]{0.85}Soft-Prompts & $\checkmark$ & 21.4 & 7.0 & 10.8 & 2.8 & 64.4 & 1.9 \\
    Hard$_{\textsc{static}}$ & & 12.0 & 10.6 & 5.0 & 3.0 & 65.7 & 2.0 \\
    \rowcolor[gray]{0.85}Hard$_{\textsc{static}}$ & $\checkmark$ & 18.8 & 6.8 & 22.0 & 7.5 & 68.8 & 1.8 \\
    
Hard$_{\textsc{dynamic}}$ & & 3.2 & 5.4 & 9.5 & 12.0 & 28.3 & 2.3 \\

\rowcolor[gray]{0.85}Hard$_{\textsc{dynamic}}$ & $\checkmark$ & 11.0 &  5.6 & 16.8 & 9.3 & 39.0 & 3.6 \\

    \bottomrule
    \end{tabular}
    \caption{Proportion [\%] of writers who gained significantly \textbf{(\textit{Ga})} improved  and \textbf{(\textit{Gb})} decreased performance, compared to the use of Fine-tuning baseline. We exploit macro-F$_1$ for Yelp and Amazon, nDCG@5 for Twitter dataset. Hard refers to Hard-Prompts.}
    \label{tab:consistency}
\end{table}

\begin{table*}[t]
    \small
    \centering
    \begin{tabular}{l@{\,\,\,\,}c@{\,\,\,\,}c@{\,\,\,\,}c@{\,\,\,\,}c@{\,\,\,\,}c@{\,\,\,\,}c@{\,\,\,\,}c@{\,\,\,\,}c@{\,\,\,\,}c@{\,\,\,\,}r@{\,\,\,\,}r@{\,\,\,\,\,}r}
    \toprule
    \multirow{2}{*}{\textbf{BERT-base}} & \multirow{2}{*}{\textbf{{Inter}}} & \multicolumn{3}{c}{\textbf{{Yelp}}} & & \multicolumn{3}{c}{\textbf{{Amazon}}} & & \multicolumn{3}{c}{\textbf{{Twitter}}}\\
    \cline{3-5}\cline{7-9}\cline{11-13}
    & & \scriptsize{\textbf{{w/o prompts}}} &  \scriptsize{\textbf{{zero-shot}}} & \scriptsize{\textbf{{approx.}}} & & \scriptsize{\textbf{{w/o prompts}}} &  \scriptsize{\textbf{{zero-shot}}} & \scriptsize{\textbf{{approx.}}} & & \scriptsize{\textbf{{w/o prompts}}} & \scriptsize{\textbf{zero-shot}} & \scriptsize{\textbf{approx.}} \\ 
    \midrule
    UserIdentifier & & 74.11 & 72.09 & 73.28 & & 72.19 & 71.85 & 72.01 & & 22.66 & 39.21 & 46.45 \\
    \rowcolor[gray]{0.85}UserIdentifier & $\checkmark$ & 75.46 & 75.63 & 75.60 & & 73.71 & 74.01 & 73.81 & &32.54 & 34.93 & 44.83 \\
    UserAdapter & & 74.36 & 73.74 & 73.73 & & 73.43 & 72.22 & 72.21 & & 38.83 & 40.91 & 44.38 \\
    \rowcolor[gray]{0.85}UserAdapter & $\checkmark$ & 76.35 & 75.89 & 75.90 & & 73.11 & 73.94 & 73.94 & & 32.35 & 36.63 & 44.96 \\
    Soft-Prompts$_{\textsc{fix}}$ & & 16.95 & 18.55 & 71.54 & & 18.55 & 18.42 & 69.96 & & 2.03 & 1.70 & 16.75 \\ 
    \rowcolor[gray]{0.85}Soft-Prompts$_{\textsc{fix}}$ & $\checkmark$ & 29.07 & 27.39 & 72.25 & & 26.09 & 26.67 & 70.24 & & 2.03 & 1.70 & 15.91 \\ 
    Soft-Prompts  & & 74.06 & 72.31 & 74.64 & & 72.31 & 71.95 & 72.07 & & 22.31 & 40.94 & 43.68 \\
    \rowcolor[gray]{0.85}Soft-Prompts & $\checkmark$  & 74.05 & 75.06 & 75.35 & & 73.31 & 73.74 & 74.20 & & 15.32 & 35.55 & 45.34 \\
    Hard-Prompts$_{\textsc{static}}$ & & 74.17 & 73.06 & 74.13 & & 73.06 & 72.01 & 72.72 & & 38.43 & 51.00 & 48.35 \\
    \rowcolor[gray]{0.85}Hard-Prompts$_{\textsc{static}}$ & $\checkmark$  & 75.20 & 75.48 & 75.68 & & 73.91 & 73.98 & 74.64 & & 35.90 & 51.52 & 46.57 \\
    
Hard-Prompts$_{\textsc{dynamic}}$ & & 73.62 & 73.78 & 73.78 & & 72.49 & 72.05 & 72.93 & & 48.87 & 52.34 & 50.10 \\
    
\rowcolor[gray]{0.85}Hard-Prompts$_{\textsc{dynamic}}$ & $\checkmark$ & 74.93 & 75.04 & 75.11 & & 74.60 & 74.11 & 74.37 & & 51.11 & 54.58 & 52.11 \\

    \hline
    Fine-tuning & & \multicolumn{3}{c}{74.09} & & \multicolumn{3}{c}{72.84} & & \multicolumn{3}{c}{53.06}\\
    \hline
    \end{tabular}
    \caption{Evaluation for examples of unknown writers. \textbf{w/o prompts}, \textbf{zero-shot}, and \textbf{approx.} refer to the non-use of prompts, the use of zero-shot prompts, and approximation to known writers, respectively. As for the evaluation metrics, we exploit macro-F1 for Yelp and Amazon, nDCG@5 for Twitter dataset.}
    \label{tab:unknown}
\end{table*}

\subsection{Performance Consistency among Writers}
We explore whether all writers consistently benefit from the personalized fine-tuning of PLMs.
By comparison with the use of the fine-tuning baseline, we divide writers into three groups and show their percentage, i.e., 
\textit{Ga}, groups that their performance significantly improves; 
\textit{Gb}, groups that their performance significantly decreases;
and \textit{Gc}, groups in which no significant change is observed.

Table~\ref{tab:consistency} shows that the numbers of writers affected positively or negatively depend greatly on the method.
\textit{Soft-Prompts}$_{\textsc{fix}}$ has larger \textit{Ga}.
We can speculate that its poor performance in main experiments \S~\ref{sec:experiment}) is due to its nature of having more writers negatively affected, i.e., \textit{Gb}.  
On the contrary, other high-performance methods, such as UserIdentifier and \textit{Hard-Prompts}$_{\textsc{static}}$, do not have an outstandingly large fraction of group \textit{Ga}, but \textit{Gb} is smaller. 
It indicates that their performance is based on the nature of not destroying the average performance during personalized fine-tuning.

Our Personalized Intermediate Learning (Inter) is consitently effective in both increasing \textit{Ga} and decreasing \textit{Gb} in most cases. We can see that our intermediate learning does not merely improve the performance of a specific number of writers.

\subsection{Performance for Unknown Writers}
We show the performance of each prompt-based method for the unknown writers who do not appear in the training, validation, and test datasets~(Table~\ref{tab:dataset}).
Concretely, we explore three simple scenarios to handle unknown writers.
First, as \citet{mireshghallah-etal-2022-useridentifier} tried, we give no prompt to data of unknown writers (\textbf{w/o prompts}). 
Next, we construct prompts on the fly, in which we create Hard-Prompts with the writer's text, generate random token sequences for UserIdentifier, and assign random weight values to \textit{Soft-Prompts} (\textbf{zero-shot}).
Finally, we approximate unknown writers to known writers (\textbf{approx.}). 
For each unknown writer, we find the known writer who wrote similar historical text, and use his/her personalized model instead. 
We compute similarity of the set of historical text across writers 
by encoding each text by Sentence-BERT~\cite{reimers-2019-sentence-bert}, averaging them per writer, and applying inner-product.

Table~\ref{tab:unknown} shows the results.
In sentiment analysis, personalized models with intermediate learning perform as well as or better than the fine-tuning model even without prompts. 
If the intermediate leraning has been applied to the models, zero-shot prompts also work. 
The most reasonable approach can be the approximation to the know writers.

On the other hand, hashtag recommendation shows performance degradation compared to fine-tuning in most cases, indicating their poor robustness to the unknown writers. 
The high performance of \textit{Hard-Prompts}$_{\textsc{dynamic}}$, which uses dynamic prompts that are not fixed to the writer, may support this claim. Although the approximation to known users is the best performing strategy to handle unknown users in this task as well, 
it is not as good as fine-tuning, indicating that the prompts of other writers works as noises in this task.

\section{Conclusion}
In this study, aiming at enabling pretrained language models (PLMs) to perform personalized text understanding, we proposed several writer-aware optimization methods.
Specifically, we introduced personalized fine-tuning methods based on various types of personalized prompts, and then proposed personalized intermediate learning method, which is based on masked language modeling on personal text.
Experiments on different text understanding tasks, datasets, and PLMs are conducted to verify these methods.

\section*{Acknowledgement}
We would like to give special thanks to Dr. Fangzhao Wu for discussing with us and giving us valuable comments on this study.

\section*{Limitations}
While we have certainly presented promising results with several datasets and pretrained language models, there are still some limitations. 
First, we focused on settings where we can not exploit the user-graphs, one of the data-dependent resource, in order to broaden the applicable situations of the methods.
However, as we introduced~(\S~\ref{sec:related_work}), the importance of the user-graph in personalizing PLMs has also been reported~\cite{hofmann2021dynamic}. 
Therefore, it may be possible to further improve personalization performance 
by designing the methods so that it can optionally use the user-graph features, if these are available. 
Second, we experimented with not extremely large PLMs such as BERT-base~\cite{devlin-etal-2019-bert}. It has been reported that the optimal design of prompts changes with the size of PLMs~\cite{lester-etal-2021-power}, though it's not for personalized prompts. 
Therefore, an evaluation of personalized prompts we introduced on larger PLMs is still under exploration.

\section*{Ethics Statement}
In general, personalized models may be used to survey or profile their personal information without their permissions~\cite{richards2013dangers,rangel2013overview}. We do not recommend the use of personalization techniques, including ours, for such ethically problematic purposes. In addition, even the accurate models can make mistakes. Therefore, it is necessary to fully discuss with the users the uncertainty of the personalized models so that important decisions are not made due to overconfidence to the models. In addition, the risk of disclosing inference results to the public needs to be discussed with the users, even if they agree with that, to prevent social attacks on the individuals. In addition, before discussing whether or not the personalized models would be used for appropriate and intended purposes, we strongly recommend to ensure that the data used for personalization is properly stored and secured.

Since our focus is the text understanding task, 
the minimum information required for the experiment is the input text and the output discrete labels. 
Therefore, it is impossible to generate personal information, which might be potential risks in personalized text generation. 
In common with all datasets, user names are converted to numerical IDs and are not used as features. 
Other personal information can not be used in this research.

\bibliography{acl2023}
\bibliographystyle{acl_natbib}

\end{document}